\documentclass[letterpaper]{article} 
\usepackage{aaai23}  
\usepackage{times}  
\usepackage{helvet}  
\usepackage{courier}  
\usepackage[hyphens]{url}  
\usepackage{graphicx} 
\def\UrlFont{\rm}  
\usepackage{natbib}  
\usepackage{caption} 
\usepackage{algorithm}
\usepackage{algorithmic}
\usepackage{amsmath}
\usepackage{amssymb}
\usepackage{xcolor}
\usepackage{siunitx}

\usepackage{newfloat}
\usepackage{listings}
\DeclareCaptionStyle{ruled}{labelfont=normalfont,labelsep=colon,strut=off} 
\floatstyle{ruled}
\newfloat{listing}{tb}{lst}{}
\floatname{listing}{Listing}
\title{Forecasting Soil Moisture Using Domain Inspired Temporal Graph Convolution Neural Networks To Guide Sustainable Crop Management.}

\author{
    Muneeza Azmat,\textsuperscript{\rm 1}
    Malvern Madondo,\textsuperscript{\rm 1}
    Kelsey Dipietro,\textsuperscript{\rm 2}
    Raya Horesh,\textsuperscript{\rm 1}
    Arun Bawa,\textsuperscript{\rm 3}
    Michael Jacobs,\textsuperscript{\rm4}
    Raghavan Srinivasan,\textsuperscript{\rm 3}
    Fearghal O'Donncha,\textsuperscript{\rm 1}
}
\affiliations{
    \textsuperscript{\rm 1}IBM Research \\
    \textsuperscript{\rm 2}Sandia National Lab\\
    \textsuperscript{\rm 3}Texas A\&M \\
    \textsuperscript{\rm 4}IBM Corporate Social Responsibility \\
}

\begin{document}

\maketitle

\begin{abstract}
Climate change, population growth, and associated water scarcity present unprecedented challenges for agriculture. As a result, climate-smart agriculture demands efficient resource usage to optimize crop production. This project aims to forecast soil moisture using domain knowledge and machine learning, which can then be used for crop management decisions that enable sustainable farming. 
Traditional methods for predicting hydrological response features, such as soil moisture, involve dividing fields into small response units and solving physics-based and empirical hydrological models, which require significant computational time and expertise. 
Recent work has implemented machine learning models as a tool for forecasting hydrological response features, but these models neglect a crucial component of traditional hydrological modeling that spatially close units can have vastly different hydrological responses. 
In traditional hydrological modeling, units
with similar hydrological properties are grouped together and
share model parameters regardless of their spatial proxim-
ity. 
Inspired by this domain knowledge, we have constructed a novel domain-inspired temporal graph convolution neural network. Our approach involves: 1) clustering units based on their time-varying hydrological properties, 2) constructing graph topologies for each cluster based on similarity using dynamic time warping, and 3) forecasting soil moisture for each unit using graph convolutions and a gated recurrent neural network. 
We have trained, validated, and tested our method on field-scale time series data consisting of approximately 99,000 hydrological response units spanning 40 years in a case study in northeastern United States.
Comparison with existing models for soil moisture forecast illustrates the effectiveness of using domain-inspired clustering with time series graph neural networks. 
The framework is currently being deployed as part of a pro bono social impact program that leverages technologies such as hybrid cloud and AI to enhance and scale non-profit and government organizations. The trained models are being deployed on a series of small-holding farms in central Texas.
\end{abstract}

\section{Introduction}
Machine learning has changed much of our world in the past decade. However, its impact in some of the most critical areas have been negligible. A prominent example is quantifying and forecasting ground- and surface- water availability to inform agricultural practices. While water plays a fundamental role, characterizing these resources involves distinct temporal and spatial processes that must consider historical and future precipitation volumes, surface and groundwater runoff from heterogeneous sources, evapotranspiration, and other water sinks or losses. 
The huge landmasses and high spatial and temporal variability makes it infeasible to collect sufficient density of observation data to implement IoT-backed decision support systems.

Traditionally, engineers have relied on physics-based models that represent hydrological processes as a set of partial differential equations constrained by heuristics, empirical relationships, and expert intuition. While these allow greater insight into spatial and temporal evolution of water over land, the associated complexity and uncertainty places a heavy burden on the expert user. Further, these models can not readily be deployed across different locations without a cumbersome calibration and validation effort. A prominent example in this regard is the Soil \& Water Assessment Tool \cite{gassman2007soil} that has been widely-used to simulate the quality and quantity of both surface and ground water processes, and inform agriculture, land use, and land management practices. A corresponding body of research has developed around parameterizing and evaluating these models with prominent examples being the parameter estimation toolbox (PEST) and SWAT Calibration and Uncertainty Program \cite{doherty2003ground, abbaspour2013swat}.


Precision agriculture approaches \cite{zhang2002precision} have developed over the past four decades by combining models, satellite, and sensor data to improve decision making.
Success in precision agriculture is related to how well it can be applied to assess, manage, and evaluate crop production \cite{pierce1999aspects}. 
Climate change introduces a completely different set of challenges that require drastically more granular data, and more holistic decision making to enable an environmentally and economically sustainable response.  The negative impacts of climate change are already being felt in the form of increasing temperatures, weather variability, shifting agroecosystem boundaries, invasive crops and pests, and more frequent extreme weather events \cite{calzadilla2013climate}. On farms, climate change is reducing crop yields, the nutritional quality of major cereals, and lowering livestock productivity \cite{world2016world}. 
These stressors particularly impact water constrained regions, resulting in groundwater depletion, soil erosion, and crop failures. 

Adapting to these challenges requires the adoption of climate-smart agriculture practices that minimize resource consumption and environmental impacts, while simultaneously ensuring food security for growing populations.
Globally, large farms increasingly digitize operations to enhance sustainability; small-holding farmers lack the skills and resources to leverage AI and IoT informed decision making. 
This led the World Economic Forum to posit that “agriculture and farming will be redefined within a decade with the adoption of AI-driven autonomous tools” \cite{wef_agri_2021}. However democratization of these solutions to small-holding and disadvantaged farmers requires scalable machine learning models that can be informed by publicly-available datasets and sparse low-cost sensor data. 

This paper describes a novel domain-inspired framework to forecast soil moisture. The proposed framework uses graph convolutional neural networks (GNN) to resolve complex hydrological response in a domain consisting of 3000 watersheds. 
 While previous research has explored a GNN approach to represent spatial patterns by superimposing a graph topology over the physical streamflow network, our approach instead generates the topology based on the degree of physical and hydrological similarity between individual watersheds. 
 This provides a more physically representative framework that is informed by the concept of group response units (GRUs), a well-established hydrological modeling technique, introduced by \cite{articleGRU}. 
GRU is composed of groups of hydrological response units (HRUs) that have similar hydrological characteristics and consequently have more comparable hydrological response than neighboring HRUs which might have different characteristics (e.g. crop versus livestock farming).
The proposed framework is applied to forecasts of soil moisture in a case study application in Northeastern United States.

The contributions of this paper are as follows:
\begin{itemize}
    \item We describe a novel domain-inspired temporal graph convolution neural network. Analogous to GRUs, a clustering algorithm based on dynamic time warping (DTW) clusters together HRUs with similar features regardless of their spatial proximity. For each cluster, the graph topology is extracted from a set of similarity metrics that encompass static and dynamic hydrological catchment attributes. 
    \item We present experimental results that compare models using our novel GNN framework against state of the art for time series forecasting, an LSTM model. These experiments demonstrate the increased gain from using hydrological feature information to inform prediction.
    \item Finally, we discuss further research opportunities to apply machine learning to improve agriculture management and environmental sustainability.  In particular, the potential to use the approach to inform regions with sparse sets of monitoring datasets. 

\end{itemize}

\section{Related Work}
\label{sec:related_work}

Recent advancements in machine learning have led to widespread interest amongst hydrologists and environmental scientists as a solution to address the challenges that persist with streamflow and run-off forecasting. While previous works have approached performance levels of state-of-the-art physics-based methods \cite{hsu1995artificial,kratzert2019toward,nearing2020deep}, the challenge remains whether it can generalize to finer scales and if it can perform in regions with limited training data. 

Physics- or empirical-based hydrological models are well established in the literature, with research in the space receiving significant impetus with the US Clean Water Act of 1977.  Data inputs to resolve streamflow processes include meteorological forcing and a large number of  parameters  describing  the  physical  characteristics  of  the  catchment (soil properties, initial water depth, topography, topology, runoff curve number, etc.) \cite{devia2015review}. Popular modelling systems include SWAT \cite{arnold2012swat}, MIKE SHE \cite{graham2005flexible}, WRF-Hydro \cite{lin2018spatiotemporal} and the VIC framework \cite{gao2010water}. On the SWAT model alone, there are over 4,500 peer-reviewed journal articles describing its application to different hydrology studies \cite{swat_lit_db_2021}.

More recently, extensive research efforts have focused on the potential of deep learning (DL) for hydrology studies \cite{shen2018transdisciplinary,shamshirband2020predicting}. In particular, research has focused on the potential of recurrent networks and LSTMs to resolve the complex, nonlinear, spatiotemporal relationship between meteorological forcing, soil moisture and streamflow \cite{kratzert2019toward}. In a provocative recent paper, \cite{nearing2021role} argued that there is significantly more information in large-scale hydrological data sets than hydrologists have been able to translate into theory or models. This argument for increased scientific insight and performance from machine learning rests on the assumption that large-scale data sets are available globally (over sufficient historical periods) to condition and inform on hydrological response. While significant progress on coarse-scale hydrology dataset curation has been achieved in the US \cite{newman2015development}, and Europe \cite{klingler2021lamah} this is not implemented for many other regions and does not approach the spatial resolutions required.

Many studies have proposed frameworks  to represent the spatiotemporal properties of geophsysical systems. The most widely used framework combines convolutional neural networks (CNN) with LSTM to represent both the spatial (CNN) and temporal (LSTM) dependencies within the data. This approach has been applied to a variety of geoscientific tasks such as precipitation nowcasting from rainfall radar maps \citep{Xingjian2015} and forecasting sea surface temperature from satellite-derived observations \citep{yang2017cfcc}. \cite{ElSaadani_2021} use this CNN + LSTM approach to estimate soil moisture. However, this approach requires gridded input data, and relies on spatial correlations. Our proposed approach overcomes these limitations by using graphs to handle unstructured data and by connecting nodes of the graph based on hydrological similarity rather than spatial proximity. 

	
An alternative approach aims to embed information from physics or heuristic knowledge within the network. 
Physics-informed DL is a novel approach for resolving information from physics. The philosophy behind it is to approximate the quantity of interest (e.g., governing equation variables) by a deep neural network (DNN) and embed the physical law to regularize the network. To this end, training the network is equivalent to minimization of a well-designed loss function that contains the PDE residuals and initial/boundary conditions \citep{rao2020physics}.

A further stream of related work has been started by \citet{chen2018neural}, who presented a novel approach to approximate the discrete series of layers between the input and output state by acting on the derivative of the hidden units. At each stage, the output of the network is computed using a black-box differential equation solver which evaluates the hidden unit dynamics to determine the solution with the desired accuracy. In effect, the parameters of the hidden unit dynamics are defined as a continuous function, which may provide greater memory efficiency and balancing of model cost against problem complexity. The approach aims to achieve comparable performance to existing state-of-the-art with far fewer parameters, and suggests potential advantages for time series modeling.

\section{Methods}
\subsection{Data}
\citet{leavesley1983precipitation} introduced the decomposition of watersheds into sub-areas that are assumed to be homogeneous in their hydrologic response, termed hydrologic response units (HRUs). The HRUs are characterized using topographic variables, such as elevation and slope, and geographic variables such as soil type, vegetation type and precipitation distribution. HRUs are generated by first decomposing a domain into a set of watersheds which represents the land area in which any precipitation eventually flows into the same outlet. Within sub-basins, HRUs 
are further delineated into smaller polygons, based on land use, soil attributes, and slope. 
For modelling and analysis, polygons with homogeneous hydrologic response are lumped together and resolved simultaneously. The concept of HRUs enable modelers to more effectively resolve complex issues regarding spatial variability to provide a more realistic representation of land surface processes \cite{prasad2005delineation}. 

We use data simulated by Soil and Water Assessment Tool (SWAT) \cite{gassman2007soil}. SWAT is the state of the art small watershed to river basin-scale model used to simulate the quality and quantity of surface and ground water and predict the environmental impact of land use, land management practices, and climate change. SWAT is widely used in developing agricultural management practices, assessing soil erosion prevention and control, non-point source pollution control and regional management in watersheds. 
While publicly available soil moisture reanalysis are available from institutions such as ECMWF (ERA Land-5) and NOAA (NLDAS), practical applications for agriculture management are constrained by the available resolution of \SI{9}{km} \cite{munoz2021era5} and \SI{14}{km} \cite{xia2012continental}, respectively. 
Agriculture, on the other hand, requires predictions that resolve field-scale ($<\SI{500}{m}$) processes.
\begin{figure}
    \centering
    \includegraphics[width=0.4\textwidth]{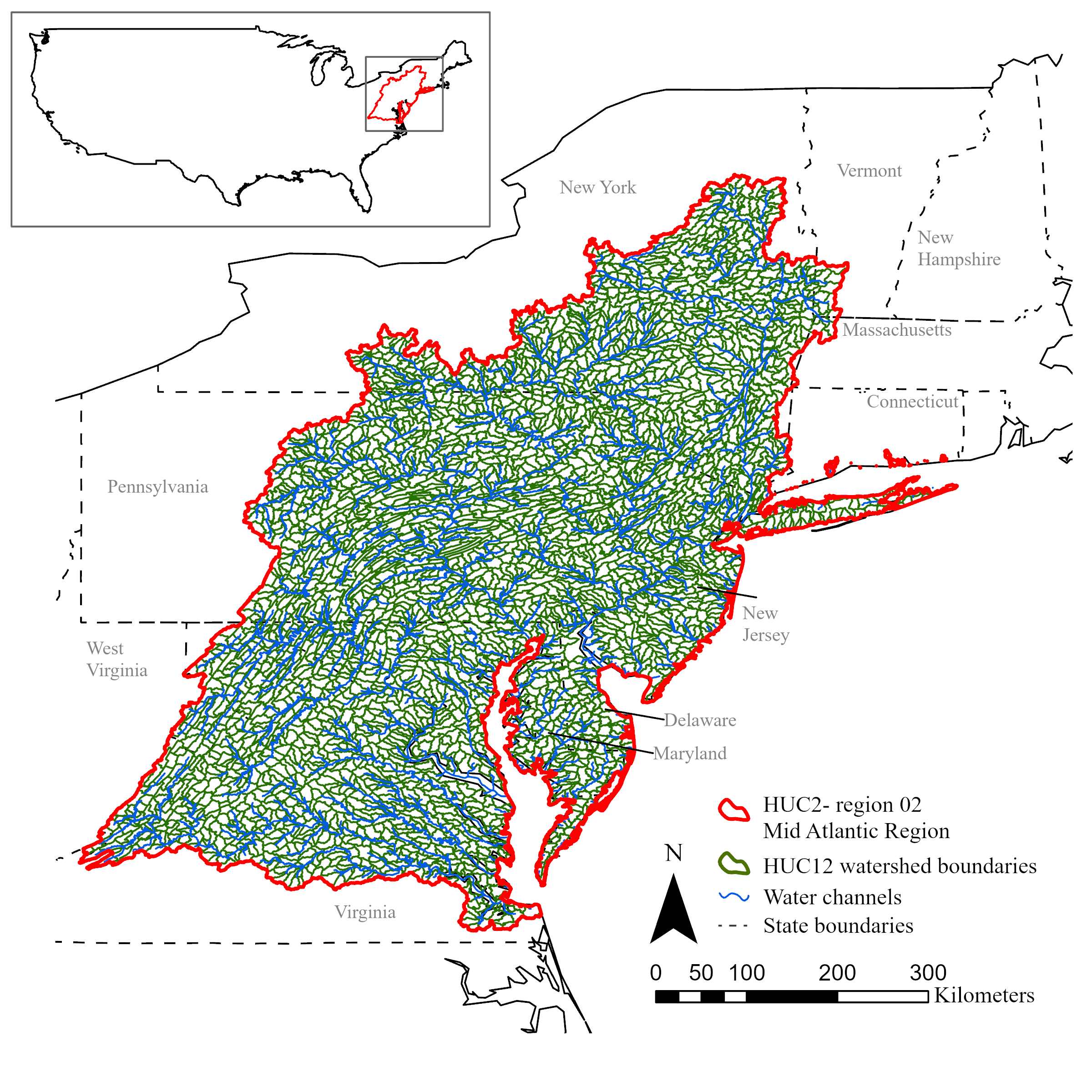}
    \caption{Layout of the Mid-Atlantic basin along with its stream network and HUC12 watersheds.}
    \label{fig:swat_layout}
\end{figure}

The Hydrological and Water Quality System (HAWQS) v2.0 \cite{chen2020analysis} (https://hawqs.tamu.edu/), a web-based interface of the SWAT model, was used to develop SWAT models for 3,037 watersheds at HUC12 (hydrologic unit code) resolution within HUC2- region 02, Mid-Atlantic region. The HAWQS provides a SWAT watershed model development framework with pre-loaded input data and modeling support capabilities for setting up models, running simulations, and processing outputs.To further divide delineated watersheds into HRUs, an area threshold of 0.5 $km^2$ was applied i.e., HRUs having area less than threshold value were not assigned a separate HRU-ID and merged with nearby HRUs. Overall, our data set consists of 3,037 watersheds divided into more than 99k HRUs. 
Detailed information about the features associated with each HRU is included in the supplementary materials. Monthly data is available for each feature spanning 34 years. 

\subsection{Problem Formulation}
Given a feature matrix $X_t \in \mathbb{R}^{n \times d}$ which is a snapshot of $d$ feature values for $n$ HRUs at time $t$, our goal is to forecast $M$ soil moisture values $\{Y_{t+i}\}_{i=0}^{M}$ in the future. 
For $M=1$ it is called single step forecasting, for $M > 1$ it is called a multi-step forecasting. We start by solving the single step forecasting problem and extend our method to multi-step forecasting.
\subsubsection{Single Forecast}
Given $X_t$ we want to forecast the soil moisture $Y_{t}$ for the next month.
\subsubsection{Multi-step Forecast}
Given $X_t$ we want to forecast the soil moisture $Y_{t} , .. , Y_{t+12}$ for the next $12$ months.

\subsection{Domain Inspired Clustering}
Inspired by the concept of group response units (GRUs), we build a clustering module to group HRUs that have similar hydrological characteristics. Traditionally, GRUs are constructed based on climate, land use, soil and pedotransfer properties \cite{HRU_clustering_Poblete2020}. The use of GRUs reduces the need for model calibration and allows for the transfer of model parameters among HRUs in the same group. 

We propose a dynamic time warping based temporal clustering technique, which leverages the seasonality of these hydrological features to inform clustering. 

First introduced in \cite{1163055, SakoeChiba71}, dynamic time warping is an algorithm for measuring the similarity between two discrete temporal signals. 
For the data tensor $\mathcal{X} \in \mathbb{R}^{n \times s \times d}$ containing $n$ HRUs , $s$ timesteps, and $d$ features $\textbf{x}^{i,j} := \mathcal{X}{i,:,j} $ represents the 1D time series data for $j$th feature in $i$th HRU. The distance matrix $D \in \mathbb{R}^{n \times n}$ represents the pairwise DTW distance between all HRUs. The distance $D_{\text{p,q}}$ between two HRUs p and q is given by
\begin{equation}
     D_{\text{p,q}} =  \sum_{j=1}^{d} DTW(\textbf{x}^{\text{p},j},\textbf{x}^{\text{q},j})  
\end{equation}
where $DTW(.,.)$ is calculated using Algorithm \ref{alg:alg_dtw}.

\begin{algorithm}[tb]
\caption{Dynamic Time Warping Algorithm}
\label{alg:alg_dtw}
\textbf{Input}: Discrete time series $\textbf{x}, \textbf{y} \in \mathbb{R}^{1 \times s}$\\
\textbf{Output}: Distance between $\textbf{x}$ and $\textbf{y}$ 
\begin{algorithmic}[1] 
\STATE initialize $C =inf \in \mathbb{R}^{n \times n}$ 
\STATE $C_{0,0}=0$
\FOR{i : $0 \rightarrow s$}
\FOR{j : $0 \rightarrow s$}
\STATE $dist = d(x_i, y_j)^2$
\STATE $C_{i,j} = dist + min(C_{i-1, j}, C_{i, j-1}, C_{i-1, j-1})$
\ENDFOR
\ENDFOR
\STATE $DTW(\textbf{x},\textbf{y}) = \sqrt{C_{s,s}}$
\STATE \textbf{return} $DTW(\textbf{x},\textbf{y})$
\end{algorithmic}
\end{algorithm}

\subsection{Temporal Graph Convolution Neural Network (TGCN)}
Graph convolution neural networks \cite{DBLP:journals/corr/KipfW16} are an extension of convolution neural networks to unstructured graph data. 
A graph $\mathcal{G}:(\mathcal{V},\mathcal{E})$ , has associated with it a set of nodes $\mathcal{V}$ connected by a set of edges $\mathcal{E}$. For our application each HRU represents a graph node. The adjacency matrix $A$ is a matrix representation of the graph topology. 

\begin{figure}[hbtp!]
    \centering
    \includegraphics[width=0.3\textwidth]{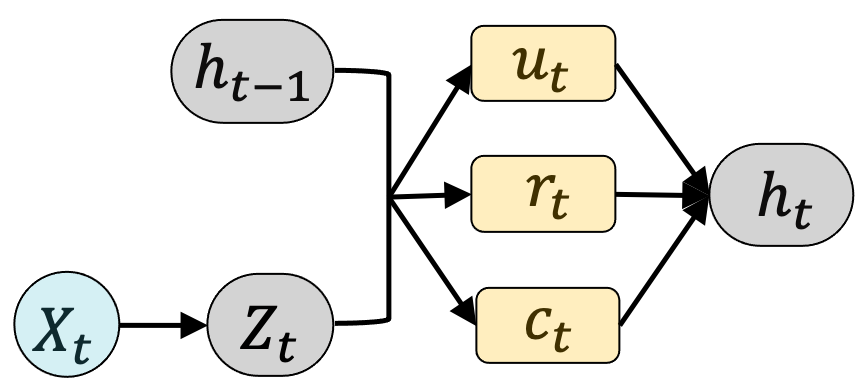}
    \caption{Schematic of a single cell of TGCN, equations (2-6).}
    \label{fig:tgcn_cell}
\end{figure}

We use the temporal graph convolution neural network detailed in \cite{8809901} for predicting soil moisture at each node. 
At time $t$, the feature matrix $X_t$ is updated using the graph convolution defined in \cite{DBLP:journals/corr/BrunaZSL13}. The resulting \textit{'neighbor-aware'} feature matrix $Z_t$ is then passed on to the gated recurrent unit (GRU). 
\begin{align}
    Z_t &= Relu (A X_t W_0 )\\
    u_t &= \sigma \left( W_u [Z_t : h_{t-1}]+ b_u\right)\\
    r_t &= \sigma \left( W_r [Z_t : h_{t-1}]+ b_r\right)\\
    c_t &= tanh \left( W_c \left[Z_t (r \odot h_{t-1}) \right] + b_c \right)\\
    h_t & = \left( u_t \odot h_{t-1}\right) + (1-u_t)\odot c_t 
\end{align}
where $u_t$ represents update gate, $r_t$ represents reset gate, $c_t$ represents cell state, $h_t$ represents hidden state, and $W_i, b_i$ are learnable weights and baises. The prediction $\hat{Y}_t$ is expressed as a linear transform of $h_t$. Figure \ref{fig:tgcn_cell} describes the information flow of a single cell of the TGCN. 

\begin{figure*}[hbtp!]
    \centering
    \includegraphics[width=0.65\textwidth]{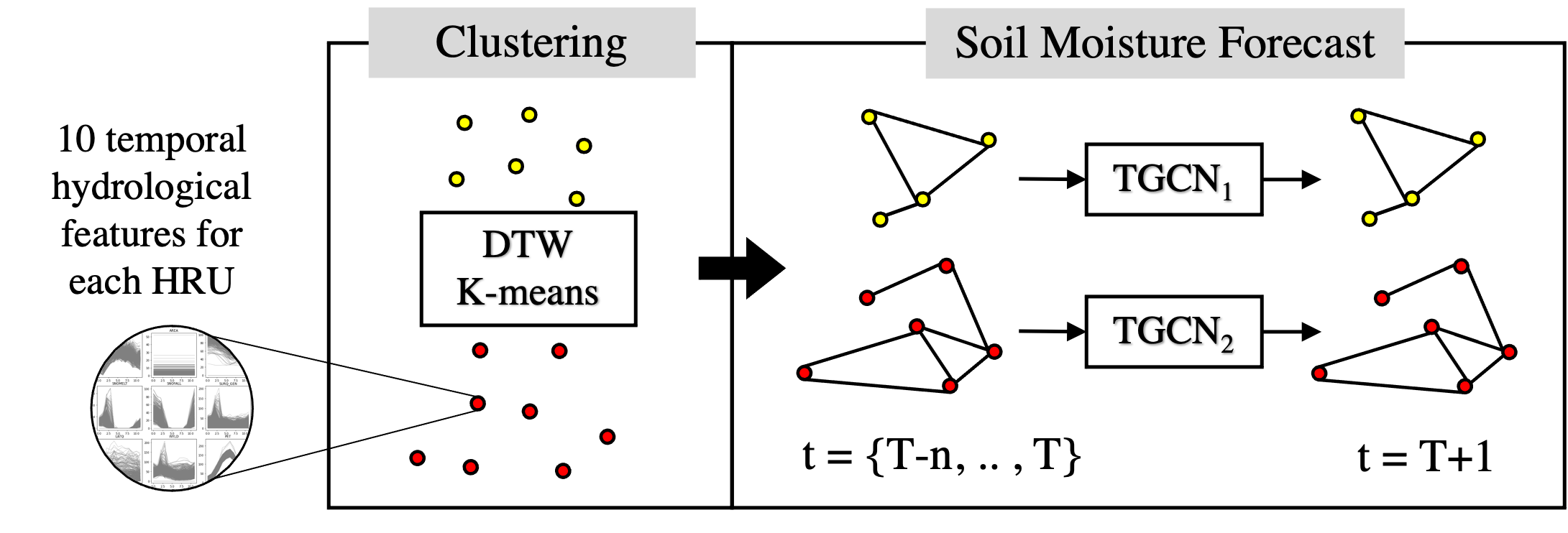}
    \caption{Schematic of our Clustering and Temporal Graph Convolution Neural network (C+TGCN) approach for soil moisture forecast.}
    \label{fig:schematic}
\end{figure*}

We minimize the mean squared error loss during training. 
\begin{equation}
    \mathcal{L}_{t} = \frac{1}{n} \sum_{i=1}^{n} (Y_{t,i} - \hat{Y}_{t,i})^2 
\end{equation}

Figure \ref{fig:schematic} summarizes the model architecture that groups similar HRUs and implements a TGCN prediction framework. 
\section{Results}

We train the LSTM model and 10 TGCN models (one for each cluster) for both the single forecast and multi-step forecasting. The number of clusters was selected based on the proportion of variance explained as described in the supplementary material.

\subsection{Evaluation Metrics}
We evaluate all models using mean squared error (MSE), which is a popular metric for regression. We also calculate the relative percent decrease in MSE to compare model performance. 

We also use Kling-Gupta Efficiency (KGE) to quantify the goodness of fit. KGE is a traditional metric used in hydrology to evaluate model performances. 
\begin{equation}
    KGE = 1- \sqrt{(r-1)^2 + (\beta -1)^2 + (\alpha -1)^2}
\end{equation}
where $r$ is the Pearson product-moment correlation coefficient, $\alpha$ is the ratio between the standard deviation of the predicted values and the standard deviation of the true values, and $\beta$ is the ratio between the mean of the predicted values and the mean of the true values. A value of $KGE = -0.41$ corresponds to using the mean value as a benchmark predictor, therefore $KGE > -0.41$  indicates that the model improves upon the mean value benchmark \cite{Knoben_2019}. As model becomes more accurate, $KGE \rightarrow 1$ . 

For model comparison, we perform a t-test to examine the statistical significance of performance improvement. Since we report test performance on independent clusters instead of k-folds, we do not violate the independence of sample assumption for the t-test. 
The null hypotheses $\mathcal{H}$ states that TGCN MSE has identical average values as LSTM MSE. For probability less than $0.05$, we reject the null hypothesis. 

\subsection{Model Details}
We use the first 27 years of data for training and validation and keep data from the last 7 years for testing. 

Before computing the DTW distance matrix $(D)$, we normalize the data using a custom min-max scaling. Instead of independently scaling the time series data, we normalize the time series for each feature by the minimum and maximum feature values across time series across all HRUs. Using this custom scaling we are able to preserve the relative trends in features. 

We use an elbow test to estimate the number of clusters. Based on the results from the elbow test we then use K-means to split HRUs into $10$ clusters. In order to avoid data leakage, we only use training data for clustering. We use functions from tslearn \cite{JMLR:v21:20-091} to implement temporal clustering.

Once the HRUs are split into clusters, we introduce graph topology on each cluster by using the DTW distances of HRUs within the cluster. The static graphs represent disjoint subsets of HRUs and are trained independently using TGCN with same model architecture. 

The model consists of a layer of graph convolution, followed by a linear transform. Output from the linear layer is then fed to the GRU which outputs the forecast $Y_t$. For multi-step model the GRU outputs a sequence of $12$ predictions for each node. All TGCNs were trained using the Adam optimizer \cite{DBLP:journals/corr/KingmaB14} with a learning rate of 1e-2 for around $100$ epochs (until validation loss stopped decreasing). Weights were initialized using He initialization \cite{7410480}. Based on the size of the graph the training took between 1.5 - 150 sec/epoch on 1 cpu core with 100G memory. Code for model setup, training and evaluation will be open-sourced on GitHub post-blind review.

\subsection{Soil Moisture Forecast Results}
\begin{figure}
    \centering
    \includegraphics[width=0.46\textwidth]{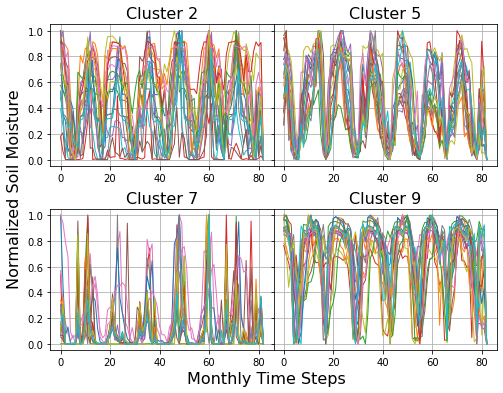}
    \caption{Plot of true soil moisture values of 20 randomly sub-sampled HRUs in cluster 2,5,7, and 9 for time steps in the test set. Soil moisture in different clusters exhibits distinct seasonal trend. }
    \label{fig:cluster_res}
\end{figure}

Figure \ref{fig:cluster_res} presents results from DTW + K-means clustering which illustrates the true normalized soil moisture values for a subset of clusters. The clusters represent distinct seasonal trends in soil moisture values exhibiting clear heterogeneity \textit{between} clusters, but high degree of similarity \textit{within} clusters. 

\begin{figure}[hbtp!]
    \centering
    \includegraphics[width=0.45\textwidth]{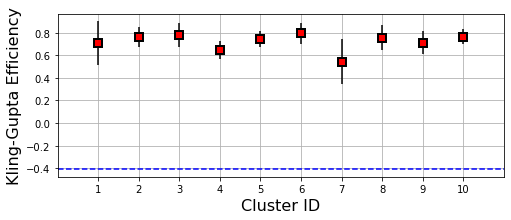}
    \caption{The plot shows the mean and standard deviation of Kling-Gupta Efficiency for each cluster for the single forecast. $KGE >-0.41$ shows that the TGCN models improve upon the mean benchmark. }
    \label{fig:kge_single}
\end{figure}

\begin{figure}[h!]
    \centering
    \includegraphics[width=0.45\textwidth]{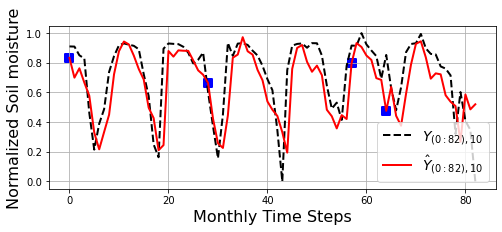}
    \caption{Predicted vs true soil moisture value for single forecast for a sample HRU(id=10) in cluster 10 from test data set. $Y$ represents ground truth, $\hat{Y}$ represents predicted soil moisture and blue boxes represent randomly sampled time steps for which multi-step results are plotted in Figure \ref{fig:multi_ypred}. }
    \label{fig:sing_forecast_pred}
    \centering
    \includegraphics[width=0.45\textwidth]{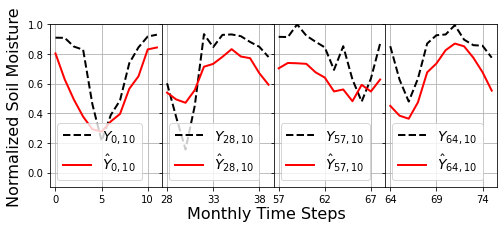}
    \caption{Predicted $(\hat{Y})$ v.s. true $Y$ soil moisture for multi-step forecast for a sample HRU(id=10) in cluster 10 from test data set.}
    \label{fig:multi_ypred}
\end{figure}

\begin{figure}[h!]
    \centering
    \includegraphics[width=0.45\textwidth]{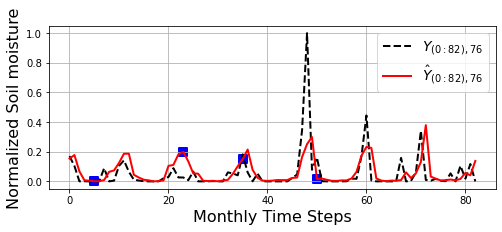}
    \caption{Predicted vs true soil moisture value for single forecast for a sample HRU(id=76) in cluster 7 from test data set. $Y$ represents ground truth, $\hat{Y}$ represents predicted soil moisture and blue boxes represent randomly sampled time steps for which multi-step results are plotted in Figure \ref{fig:multi_ypred2}. }
    \label{fig:sing_forecast_pred2}
    \centering
    \includegraphics[width=0.45\textwidth]{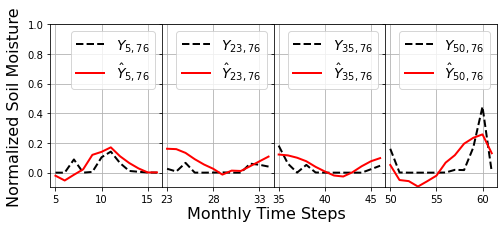}
    \caption{Predicted $(\hat{Y})$ v.s. true $Y$ soil moisture for multi-step forecast for a sample HRU(id=76) in cluster 7 from test data set. }
    \label{fig:multi_ypred2}
\end{figure}

Table \ref{MSE_singlestep} shows the average mean squared error of predicted soil moisture in each cluster. The proposed framework reduces the mean squared error across all clusters compared to the LSTM model. Figure \ref{fig:kge_single} shows the mean and standard deviation of KGE for all HRUs in a cluster. Each region reports KGE between 0.4--0.7 indicating an effective model. All instances report values greater than $-0.41$ illustrating clear improvement upon a naive model. Figure \ref{fig:sing_forecast_pred} and \ref{fig:sing_forecast_pred2} show predicted values of soil moisture on a sample HRU compared to the true values. These figures illustrate the logic behind our approach, where the TGCN corresponding to every cluster is being trained to predict different trends in soil moisture, analogous to GRUs sharing model parameters in traditional hydrological modeling.

\begin{figure}[hbtp]
    \centering
    \includegraphics[width=0.45\textwidth]{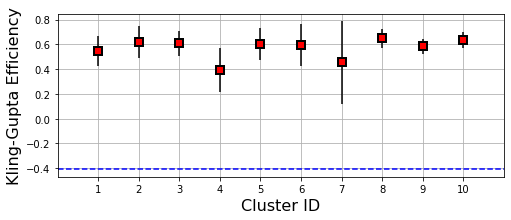}
    \caption{The plot shows the mean and standard deviation of Kling-Gupta Efficiency for each cluster for the multi-step forecast. $KGE >-0.41$ shows that the TGCN models improve upon the mean benchmark. }
    \label{fig:kge_mult}
\end{figure}

Table \ref{MSE_Multistep} shows a comparison of the prediction mean squared error for multi-step forecasting. The average mean squared error of LSTM across all clusters is  0.4584 with a standard deviation of 0.2179. Whereas, the average mean squared error of our method across all clusters is  0.0480 with a standard deviation of 0.0165. The p value of null-hypothesis $\mathcal{H}$ is 6.5e-6, which shows that the reduction in predicted mean squared error of our model compared to LSTM is statistically significant. Figure \ref{fig:kge_mult} shows that on average our method improves upon a naive model. 

\begin{table}[hbtp!]
    \centering
    \begin{tabular}{cccc}
\hline
\multicolumn{1}{|c|}{\textbf{\begin{tabular}[c]{@{}c@{}}Cluster \\ ID\end{tabular}}} &
  \multicolumn{1}{c|}{\textbf{\begin{tabular}[c]{@{}c@{}}LSTM \\ MSE\end{tabular}}} &
  \multicolumn{1}{c|}{\textbf{\begin{tabular}[c]{@{}c@{}}C+TGCN \\MSE\end{tabular}}} &
  \multicolumn{1}{c|}{\textbf{\begin{tabular}[c]{@{}c@{}}Relative\\ MSE \\Reduction\end{tabular}}} \\ \hline
\multicolumn{1}{|c|}{1}  & \multicolumn{1}{c|}{0.3433} & \multicolumn{1}{c|}{0.0332} & \multicolumn{1}{c|}{90.34\%} \\ \hline
\multicolumn{1}{|c|}{2}  & \multicolumn{1}{c|}{0.3815} & \multicolumn{1}{c|}{0.0328} & \multicolumn{1}{c|}{91.41\%} \\ \hline
\multicolumn{1}{|c|}{3}  & \multicolumn{1}{c|}{0.3588} & \multicolumn{1}{c|}{0.0283} & \multicolumn{1}{c|}{92.12\%} \\ \hline
\multicolumn{1}{|c|}{4}  & \multicolumn{1}{c|}{0.3057} & \multicolumn{1}{c|}{0.0399} & \multicolumn{1}{c|}{86.95\%}  \\ \hline
\multicolumn{1}{|c|}{5}  & \multicolumn{1}{c|}{0.3677} & \multicolumn{1}{c|}{0.0307} & \multicolumn{1}{c|}{91.64\%} \\ \hline
\multicolumn{1}{|c|}{6}  & \multicolumn{1}{c|}{0.4087} & \multicolumn{1}{c|}{0.0321} & \multicolumn{1}{c|}{92.14\%} \\ \hline
\multicolumn{1}{|c|}{7}  & \multicolumn{1}{c|}{0.7326} & \multicolumn{1}{c|}{0.0389} & \multicolumn{1}{c|}{94.69\%} \\ \hline
\multicolumn{1}{|c|}{8}  & \multicolumn{1}{c|}{0.4010} & \multicolumn{1}{c|}{0.0217} & \multicolumn{1}{c|}{94.58\%} \\ \hline
\multicolumn{1}{|c|}{9}  & \multicolumn{1}{c|}{0.4227} & \multicolumn{1}{c|}{0.0383} & \multicolumn{1}{c|}{90.93\%} \\ \hline
\multicolumn{1}{|c|}{10} & \multicolumn{1}{c|}{0.3847} & \multicolumn{1}{c|}{0.0335} & \multicolumn{1}{c|}{91.30\%} \\ \hline
\multicolumn{1}{l}{}     & \multicolumn{1}{l}{}        & \multicolumn{1}{l}{}        & \multicolumn{1}{l}{}        
\end{tabular}
    \caption{Mean Squared Error (MSE) for single soil moisture forecast across clusters using TGCN, compared with LSTM model.}
    \label{MSE_singlestep}
\end{table}

\begin{table}[hbtp!]
\centering
\begin{tabular}{ccccc}
\hline
\multicolumn{1}{|c|}{\textbf{\begin{tabular}[c]{@{}c@{}}Cluster \\ ID\end{tabular}}} &
  \multicolumn{1}{c|}{\textbf{\begin{tabular}[c]{@{}c@{}}LSTM \\ MSE\end{tabular}}} &
  \multicolumn{1}{c|}{\textbf{\begin{tabular}[c]{@{}c@{}}C+TGCN \\ MSE\end{tabular}}} &
  \multicolumn{1}{c|}{\textbf{\begin{tabular}[c]{@{}c@{}}Relative\\ MSE \\Reduction\end{tabular}}} \\ \hline
\multicolumn{1}{|c|}{1}   & \multicolumn{1}{c|}{0.3433} & \multicolumn{1}{c|}{0.0549} & \multicolumn{1}{c|}{82.93\%} \\ \hline
\multicolumn{1}{|c|}{2}  & \multicolumn{1}{c|}{0.3815} & \multicolumn{1}{c|}{0.0573} & \multicolumn{1}{c|}{84.93\%} \\ \hline
\multicolumn{1}{|c|}{3}  & \multicolumn{1}{c|}{0.3588} & \multicolumn{1}{c|}{0.0523} & \multicolumn{1}{c|}{86.06\%} \\ \hline
\multicolumn{1}{|c|}{4}   & \multicolumn{1}{c|}{0.3057}& \multicolumn{1}{c|}{0.0610} & \multicolumn{1}{c|}{79.60\%} \\ \hline
\multicolumn{1}{|c|}{5}   & \multicolumn{1}{c|}{0.3677}& \multicolumn{1}{c|}{0.0527} & \multicolumn{1}{c|}{86.06\%} \\ \hline
\multicolumn{1}{|c|}{6}   & \multicolumn{1}{c|}{0.4087}& \multicolumn{1}{c|}{0.0543} & \multicolumn{1}{c|}{86.19\%} \\ \hline
\multicolumn{1}{|c|}{7}   & \multicolumn{1}{c|}{0.7326}& \multicolumn{1}{c|}{0.0417} & \multicolumn{1}{c|}{94.29\%} \\ \hline
\multicolumn{1}{|c|}{8}   & \multicolumn{1}{c|}{0.4010}& \multicolumn{1}{c|}{0.0393} & \multicolumn{1}{c|}{91.10\%} \\ \hline
\multicolumn{1}{|c|}{9}   & \multicolumn{1}{c|}{0.4227}& \multicolumn{1}{c|}{0.0560} & \multicolumn{1}{c|}{87.42\%} \\ \hline
\multicolumn{1}{|c|}{10}  & \multicolumn{1}{c|}{0.3847}& \multicolumn{1}{c|}{0.0591} & \multicolumn{1}{c|}{83.43\%} \\ \hline
\multicolumn{1}{l}{}     & \multicolumn{1}{l}{}        & \multicolumn{1}{l}{}        & \multicolumn{1}{l}{}      
\end{tabular}
\caption{Mean Squared Error (MSE) for multi-step soil moisture forecast across clusters using Clustering and TGCN (C+TGCN), compared with the LSTM model.}
\label{MSE_Multistep}
\end{table}

\section{Discussion}
Soil moisture estimation and prediction are critical to climate-aware agriculture. Resolving these predictions requires a comprehensive assessment of heterogeneous spatial and temporal features. While well-established physics-based approaches exist, they are hindered by their high user complexity and computational expense to deploy at scale for commodity use cases. In practical terms, they are the domain of academic institutions and government organizations.

This paper describes a machine learning framework that borrows concepts from hydrological modelling to improve predictive skill and ease interpretability.
There is a large volume of literature related to applying physics-informed constraints to ML which were discussed earlier. The objective in many of those studies is to augment the models with data external to the training data via methods such as modified loss functions \cite{daw2020physics}, data augmentation \cite{James2018}, or specifying consensus filters to guide disparate models or data towards convergence \citep{haehnel2020using}. 


The proposed methodology presents a natural framework to ingest information external to a time series signal positing the opportunity to enhance learning.
Results demonstrate large increase in predictive skill provided by the GNN framework. Conventional techniques such as LSTM are commonly used for soil moisture prediction \citep{li2022attention}. However, these approaches treat different locations independently and fail to exploit spatial dependencies.

\citet{vyas2020dynamic} described a GNN approach to forecast soil moisture based on Dynamic Graph Learning. At each timestep graph topology is updated based on a smoothness regularizer that evaluated dissimilarity for both node features and labels. Regularized dynamic graph updating have demonstrated improved model prediction in general cases \cite{chen2020measuring}. However, for soil moisture prediction, graph connectivity can be more effectively informed based on a systematic quantification of static and dynamic catchment attributes. Due to the high spatial and temporal hetereogeneity dynamic updating can lead to spurious correlations based on synoptic similarity between features or labels.
This is exacerbated by the long heterogeneous memory of soil moisture concentration. For example, the soil moisture at a point depends on weather processes together with previous moisture values over a specific window. The length of the historic window is highly dependent on local factors such as soil types, vegetation cover, and slope. For example, clay soils will have longer moisture retention than sandy soils. 
To accurately represent these dynamics, graph topology need to consider hydrological processes and their implications rather than individual physical descriptors.

A prominent body of literature has explored the combination of CNN and LSTM frameworks to resolve spatiotemporal processes. (e.g. \cite{Xingjian2015, yang2017cfcc}). These provide an intuitive and pragmatic approach to incorporate these information dimensions but are generally constrained to data on a consistent spatial grid. Applications have exclusively focused on gridded data such as satellite measurements, radar observations, and numerical model reanalysis products. 
Our proposed GNN framework adapts naturally to the characteristics of hydrological data. Individual polygons or hydrological response units are characterised based on their specific properties and informs a message passing between different regions based on similarity. 
Further, our approach provides a direct fit to modern Internet of Things (IoT) sensor networks that are typically of limited spatial dimension but have complex (often time-lagged) dependencies between neighboring sensors. With information on the hydrological features, a graph topology can be constructed connecting different sensors informed by established physics-based relationships.

\section{Conclusion}
This paper describes a spatiotemporal soil moisture prediction framework. Robust, high-resolution estimates are critical to most aspects of farm management, including: planting and harvesting scheduling, drought and irrigation management, and informing insurance risk and coverage.
Creating a graph topology based on similarity metrics rather than the physical stream network and topography improved prediction performance by 70--90\%.
Further, decoupling the graph topology from spatial relationships improves the generalizability of the framework. The approach can be applied to regions that share properties such as climate, soil features, and vegetation, regardless of spatial proximity. This has the attractive property that data from regions with well developed monitoring programs can inform predictions in other locations or geographies. 
Estimating and forecasting soil moisture in ungauged basins is one of the great challenges of hydrology.  
This implicit form of parameters sharing enabled by the spatially decoupled graph network is a valuable contribution to this ambition. Informed by well-established hydrological understanding and using a computationally efficient TGCN, the framework is particularly applicable for regions with limited computational resources or observation data. This is particularly important in hydrology where collecting high-quality data is time consuming and expensive.


\bibliography{aaai23}

\begin{thebibliography}{49}
\providecommand{\natexlab}[1]{#1}

\bibitem[{Abbaspour(2013)}]{abbaspour2013swat}
Abbaspour, K.~C. 2013.
\newblock Swat-cup 2012.
\newblock \emph{SWAT calibration and uncertainty program—A user manual}.

\bibitem[{Arnold et~al.(2012)Arnold, Moriasi, Gassman, Abbaspour, White,
  Srinivasan, Santhi, Harmel, Van~Griensven, Van~Liew et~al.}]{arnold2012swat}
Arnold, J.~G.; Moriasi, D.~N.; Gassman, P.~W.; Abbaspour, K.~C.; White, M.~J.;
  Srinivasan, R.; Santhi, C.; Harmel, R.; Van~Griensven, A.; Van~Liew, M.~W.;
  et~al. 2012.
\newblock SWAT: Model use, calibration, and validation.
\newblock \emph{Transactions of the ASABE}, 55(4): 1491--1508.

\bibitem[{Bank(2016)}]{world2016world}
Bank, W. 2016.
\newblock World Bank group climate change action plan.

\bibitem[{Bruna et~al.(2014)Bruna, Zaremba, Szlam, and
  LeCun}]{DBLP:journals/corr/BrunaZSL13}
Bruna, J.; Zaremba, W.; Szlam, A.; and LeCun, Y. 2014.
\newblock Spectral Networks and Locally Connected Networks on Graphs.
\newblock In Bengio, Y.; and LeCun, Y., eds., \emph{2nd International
  Conference on Learning Representations, {ICLR} 2014, Banff, AB, Canada, April
  14-16, 2014, Conference Track Proceedings}.

\bibitem[{Calzadilla et~al.(2013)Calzadilla, Rehdanz, Betts, Falloon,
  Wiltshire, and Tol}]{calzadilla2013climate}
Calzadilla, A.; Rehdanz, K.; Betts, R.; Falloon, P.; Wiltshire, A.; and Tol,
  R.~S. 2013.
\newblock Climate change impacts on global agriculture.
\newblock \emph{Climatic change}, 120(1): 357--374.

\bibitem[{Chen et~al.(2020{\natexlab{a}})Chen, Lin, Li, Li, Zhou, and
  Sun}]{chen2020measuring}
Chen, D.; Lin, Y.; Li, W.; Li, P.; Zhou, J.; and Sun, X. 2020{\natexlab{a}}.
\newblock Measuring and relieving the over-smoothing problem for graph neural
  networks from the topological view.
\newblock In \emph{Proceedings of the AAAI Conference on Artificial
  Intelligence}, volume~34, 3438--3445.

\bibitem[{Chen et~al.(2020{\natexlab{b}})Chen, Gassman, Srinivasan, Cui, and
  Arritt}]{chen2020analysis}
Chen, M.; Gassman, P.~W.; Srinivasan, R.; Cui, Y.; and Arritt, R.
  2020{\natexlab{b}}.
\newblock Analysis of alternative climate datasets and evapotranspiration
  methods for the Upper Mississippi River Basin using SWAT within HAWQS.
\newblock \emph{Science of the Total Environment}, 720: 137562.

\bibitem[{Chen et~al.(2018)Chen, Rubanova, Bettencourt, and
  Duvenaud}]{chen2018neural}
Chen, R.~T.; Rubanova, Y.; Bettencourt, J.; and Duvenaud, D.~K. 2018.
\newblock Neural ordinary differential equations.
\newblock In \emph{Advances in neural information processing systems},
  volume~31.

\bibitem[{Daw et~al.(2020)Daw, Thomas, Carey, Read, Appling, and
  Karpatne}]{daw2020physics}
Daw, A.; Thomas, R.~Q.; Carey, C.~C.; Read, J.~S.; Appling, A.~P.; and
  Karpatne, A. 2020.
\newblock Physics-guided architecture (pga) of neural networks for quantifying
  uncertainty in lake temperature modeling.
\newblock In \emph{Proceedings of the 2020 siam international conference on
  data mining}, 532--540. SIAM.

\bibitem[{Devia, Ganasri, and Dwarakish(2015)}]{devia2015review}
Devia, G.~K.; Ganasri, B.~P.; and Dwarakish, G.~S. 2015.
\newblock A review on hydrological models.
\newblock \emph{Aquatic procedia}, 4: 1001--1007.

\bibitem[{Doherty(2003)}]{doherty2003ground}
Doherty, J. 2003.
\newblock Ground water model calibration using pilot points and regularization.
\newblock \emph{Groundwater}, 41(2): 170--177.

\bibitem[{ElSaadani et~al.(2021)ElSaadani, Habib, Abdelhameed, and
  Bayoumi}]{ElSaadani_2021}
ElSaadani, M.; Habib, E.; Abdelhameed, A.~M.; and Bayoumi, M. 2021.
\newblock Assessment of a Spatiotemporal Deep Learning Approach for Soil
  Moisture Prediction and Filling the Gaps in Between Soil Moisture
  Observations.
\newblock \emph{Frontiers in Artificial Intelligence}, 4.

\bibitem[{Gao et~al.(2009)Gao, Tang, Shi, Zhu, Bohn, Su, Sheffield, Pan,
  Lettenmaier, and Wood}]{gao2010water}
Gao, H.; Tang, Q.; Shi, X.; Zhu, C.; Bohn, T.; Su, F.; Sheffield, J.; Pan, M.;
  Lettenmaier, D.; and Wood, E. 2009.
\newblock Algorithm Theoretical Basis Document “Water Budget Record from
  Variable Infiltration Capacity (VIC) Model”.
\newblock \emph{Algorithm Theoretical Basis Document for Terrestrial Water
  Cycle Data Records}.

\bibitem[{Gassman et~al.(2007)Gassman, Reyes, Green, and
  Arnold}]{gassman2007soil}
Gassman, P.~W.; Reyes, M.~R.; Green, C.~H.; and Arnold, J.~G. 2007.
\newblock The soil and water assessment tool: historical development,
  applications, and future research directions.
\newblock \emph{Transactions of the ASABE}, 50(4): 1211--1250.

\bibitem[{Graham and Butts(2005)}]{graham2005flexible}
Graham, D.~N.; and Butts, M.~B. 2005.
\newblock Flexible, integrated watershed modelling with MIKE SHE.
\newblock \emph{Watershed models}, 849336090: 245--272.

\bibitem[{Haehnel et~al.(2020)Haehnel, Mare{\v{c}}ek, Monteil, and
  O'Donncha}]{haehnel2020using}
Haehnel, P.; Mare{\v{c}}ek, J.; Monteil, J.; and O'Donncha, F. 2020.
\newblock Using deep learning to extend the range of air pollution monitoring
  and forecasting.
\newblock \emph{Journal of Computational Physics}, 408: 109278.

\bibitem[{He et~al.(2015)He, Zhang, Ren, and Sun}]{7410480}
He, K.; Zhang, X.; Ren, S.; and Sun, J. 2015.
\newblock Delving Deep into Rectifiers: Surpassing Human-Level Performance on
  ImageNet Classification.
\newblock In \emph{2015 IEEE International Conference on Computer Vision
  (ICCV)}, 1026--1034.

\bibitem[{Hsu, Gupta, and Sorooshian(1995)}]{hsu1995artificial}
Hsu, K.-l.; Gupta, H.~V.; and Sorooshian, S. 1995.
\newblock Artificial neural network modeling of the rainfall-runoff process.
\newblock \emph{Water resources research}, 31(10): 2517--2530.

\bibitem[{Itzhaky(2021)}]{wef_agri_2021}
Itzhaky, R. 2021.
\newblock How AI will solve agriculture's water efficiency problems.
\newblock
  \url{https://www.weforum.org/agenda/2021/01/ai-agriculture-water-irrigation-farming/}.

\bibitem[{James, Zhang, and O'Donncha(2018)}]{James2018}
James, S.; Zhang, Y.; and O'Donncha, F. 2018.
\newblock {A machine learning framework to forecast wave conditions}.
\newblock \emph{Coastal Engineering}, 137.

\bibitem[{Kingma and Ba(2015)}]{DBLP:journals/corr/KingmaB14}
Kingma, D.~P.; and Ba, J. 2015.
\newblock Adam: A Method for Stochastic Optimization.
\newblock In \emph{ICLR (Poster)}.

\bibitem[{Kipf and Welling(2016)}]{DBLP:journals/corr/KipfW16}
Kipf, T.~N.; and Welling, M. 2016.
\newblock Semi-Supervised Classification with Graph Convolutional Networks.
\newblock \emph{CoRR}, abs/1609.02907.

\bibitem[{Klingler, Schulz, and Herrnegger(2021)}]{klingler2021lamah}
Klingler, C.; Schulz, K.; and Herrnegger, M. 2021.
\newblock LamaH| Large-Sample Data for Hydrology and Environmental Sciences for
  Central Europe.
\newblock \emph{Earth System Science Data Discussions}, 1--46.

\bibitem[{Knoben, Freer, and Woods(2019)}]{Knoben_2019}
Knoben, W. J.~M.; Freer, J.~E.; and Woods, R.~A. 2019.
\newblock Technical note: Inherent benchmark or not? Comparing
  Nash{\textendash}Sutcliffe and Kling{\textendash}Gupta efficiency scores.
\newblock \emph{Hydrology and Earth System Sciences}, 23(10): 4323--4331.

\bibitem[{Kouwen et~al.(1993)Kouwen, Soulis, Pietroniro, JR, and
  Harrington}]{articleGRU}
Kouwen, N.; Soulis, E.; Pietroniro, A.; JR, D.; and Harrington, R. 1993.
\newblock Grouped Response Units for Distributed Hydrologic Modeling.
\newblock \emph{Journal of Water Resources Planning and Management-asce - J
  WATER RESOUR PLAN MAN-ASCE}, 119.

\bibitem[{Kratzert et~al.(2019)Kratzert, Klotz, Herrnegger, Sampson,
  Hochreiter, and Nearing}]{kratzert2019toward}
Kratzert, F.; Klotz, D.; Herrnegger, M.; Sampson, A.~K.; Hochreiter, S.; and
  Nearing, G.~S. 2019.
\newblock Toward improved predictions in ungauged basins: Exploiting the power
  of machine learning.
\newblock \emph{Water Resources Research}, 55(12): 11344--11354.

\bibitem[{Leavesley et~al.(1983)Leavesley, Lichty, Troutman, and
  Saindon}]{leavesley1983precipitation}
Leavesley, G.; Lichty, R.; Troutman, B.; and Saindon, L. 1983.
\newblock Precipitation-runoff modeling system: User’s manual.
\newblock \emph{Water-resources investigations report}, 83: 4238.

\bibitem[{Li et~al.(2022)Li, Zhu, Shangguan, Wang, Li, and
  Yu}]{li2022attention}
Li, Q.; Zhu, Y.; Shangguan, W.; Wang, X.; Li, L.; and Yu, F. 2022.
\newblock An attention-aware LSTM model for soil moisture and soil temperature
  prediction.
\newblock \emph{Geoderma}, 409: 115651.

\bibitem[{Lin et~al.(2018)Lin, Rajib, Yang, Somos-Valenzuela, Merwade,
  Maidment, Wang, and Chen}]{lin2018spatiotemporal}
Lin, P.; Rajib, M.~A.; Yang, Z.-L.; Somos-Valenzuela, M.; Merwade, V.;
  Maidment, D.~R.; Wang, Y.; and Chen, L. 2018.
\newblock Spatiotemporal evaluation of simulated evapotranspiration and
  streamflow over Texas using the WRF-Hydro-RAPID modeling framework.
\newblock \emph{JAWRA Journal of the American Water Resources Association},
  54(1): 40--54.

\bibitem[{Mu{\~n}oz-Sabater et~al.(2021)Mu{\~n}oz-Sabater, Dutra,
  Agust{\'\i}-Panareda, Albergel, Arduini, Balsamo, Boussetta, Choulga,
  Harrigan, Hersbach et~al.}]{munoz2021era5}
Mu{\~n}oz-Sabater, J.; Dutra, E.; Agust{\'\i}-Panareda, A.; Albergel, C.;
  Arduini, G.; Balsamo, G.; Boussetta, S.; Choulga, M.; Harrigan, S.; Hersbach,
  H.; et~al. 2021.
\newblock ERA5-Land: A state-of-the-art global reanalysis dataset for land
  applications.
\newblock \emph{Earth System Science Data}, 13(9): 4349--4383.

\bibitem[{Nearing et~al.(2020)Nearing, Kratzert, Klotz, Hoedt, Klambauer,
  Hochreiter, Gupta, Nevo, and Matias}]{nearing2020deep}
Nearing, G.; Kratzert, F.; Klotz, D.; Hoedt, P.-J.; Klambauer, G.; Hochreiter,
  S.; Gupta, H.; Nevo, S.; and Matias, Y. 2020.
\newblock A Deep Learning Architecture for Conservative Dynamical Systems:
  Application to Rainfall-Runoff Modeling.
\newblock \emph{AI for Earth Sciences Workshop at NeurIPS}.

\bibitem[{Nearing et~al.(2021)Nearing, Kratzert, Sampson, Pelissier, Klotz,
  Frame, Prieto, and Gupta}]{nearing2021role}
Nearing, G.~S.; Kratzert, F.; Sampson, A.~K.; Pelissier, C.~S.; Klotz, D.;
  Frame, J.~M.; Prieto, C.; and Gupta, H.~V. 2021.
\newblock What role does hydrological science play in the age of machine
  learning?
\newblock \emph{Water Resources Research}, 57(3): e2020WR028091.

\bibitem[{Newman et~al.(2015)Newman, Clark, Sampson, Wood, Hay, Bock, Viger,
  Blodgett, Brekke, Arnold et~al.}]{newman2015development}
Newman, A.; Clark, M.; Sampson, K.; Wood, A.; Hay, L.; Bock, A.; Viger, R.;
  Blodgett, D.; Brekke, L.; Arnold, J.; et~al. 2015.
\newblock Development of a large-sample watershed-scale hydrometeorological
  data set for the contiguous USA: data set characteristics and assessment of
  regional variability in hydrologic model performance.
\newblock \emph{Hydrology and Earth System Sciences}, 19(1): 209--223.

\bibitem[{Pierce and Nowak(1999)}]{pierce1999aspects}
Pierce, F.~J.; and Nowak, P. 1999.
\newblock Aspects of precision agriculture.
\newblock \emph{Advances in agronomy}, 67: 1--85.

\bibitem[{Poblete et~al.(2020)Poblete, Arevalo, Nicolis, and
  Figueroa}]{HRU_clustering_Poblete2020}
Poblete, D.; Arevalo, J.; Nicolis, O.; and Figueroa, F. 2020.
\newblock Optimization of Hydrologic Response Units ({HRUs}) Using Gridded
  Meteorological Data and Spatially Varying Parameters.
\newblock \emph{Water}, 12(12): 3558.

\bibitem[{Prasad(2005)}]{prasad2005delineation}
Prasad, V.~H. 2005.
\newblock Delineation of Hydrologic Response Units (HRUs) using Remote Sensing
  and GIS.
\newblock \emph{Water and Energy Abstracts}, 15(1).

\bibitem[{Rao, Sun, and Liu(2020)}]{rao2020physics}
Rao, C.; Sun, H.; and Liu, Y. 2020.
\newblock Physics informed deep learning for computational elastodynamics
  without labeled data.
\newblock \emph{arXiv preprint arXiv:2006.08472}.

\bibitem[{Sakoe(1971)}]{SakoeChiba71}
Sakoe, H. 1971.
\newblock Dynamic-programming approach to continuous speech recognition.
\newblock In \emph{1971 Proc. the International Congress of Acoustics,
  Budapest}.

\bibitem[{Sakoe and Chiba(1978)}]{1163055}
Sakoe, H.; and Chiba, S. 1978.
\newblock Dynamic programming algorithm optimization for spoken word
  recognition.
\newblock \emph{IEEE Transactions on Acoustics, Speech, and Signal Processing},
  26(1): 43--49.

\bibitem[{Shamshirband et~al.(2020)Shamshirband, Hashemi, Salimi, Samadianfard,
  Asadi, Shadkani, Kargar, Mosavi, Nabipour, and
  Chau}]{shamshirband2020predicting}
Shamshirband, S.; Hashemi, S.; Salimi, H.; Samadianfard, S.; Asadi, E.;
  Shadkani, S.; Kargar, K.; Mosavi, A.; Nabipour, N.; and Chau, K.-W. 2020.
\newblock Predicting standardized streamflow index for hydrological drought
  using machine learning models.
\newblock \emph{Engineering Applications of Computational Fluid Mechanics},
  14(1): 339--350.

\bibitem[{Shen(2018)}]{shen2018transdisciplinary}
Shen, C. 2018.
\newblock A transdisciplinary review of deep learning research and its
  relevance for water resources scientists.
\newblock \emph{Water Resources Research}, 54(11): 8558--8593.

\bibitem[{Srinivasan and Balmer(2021)}]{swat_lit_db_2021}
Srinivasan, R.; and Balmer, C. 2021.
\newblock {SWAT Literature Database for Peer-Reviewed Journal Articles}.

\bibitem[{Tavenard et~al.(2020)Tavenard, Faouzi, Vandewiele, Divo, Androz,
  Holtz, Payne, Yurchak, Ru{\ss}wurm, Kolar, and Woods}]{JMLR:v21:20-091}
Tavenard, R.; Faouzi, J.; Vandewiele, G.; Divo, F.; Androz, G.; Holtz, C.;
  Payne, M.; Yurchak, R.; Ru{\ss}wurm, M.; Kolar, K.; and Woods, E. 2020.
\newblock Tslearn, A Machine Learning Toolkit for Time Series Data.
\newblock \emph{Journal of Machine Learning Research}, 21(118): 1--6.

\bibitem[{Vyas and Bandyopadhyay(2022)}]{vyas2020dynamic}
Vyas, A.; and Bandyopadhyay, S. 2022.
\newblock Dynamic Structure Learning through Graph Neural Network for
  Forecasting Soil Moisture in Precision Agriculture.
\newblock In \emph{Proceedings of the 2022 International Joint Conference on
  Artificial Intelligence}, 5185--5191. IJCAI.

\bibitem[{Xia et~al.(2012)Xia, Mitchell, Ek, Sheffield, Cosgrove, Wood, Luo,
  Alonge, Wei, Meng et~al.}]{xia2012continental}
Xia, Y.; Mitchell, K.; Ek, M.; Sheffield, J.; Cosgrove, B.; Wood, E.; Luo, L.;
  Alonge, C.; Wei, H.; Meng, J.; et~al. 2012.
\newblock Continental-scale water and energy flux analysis and validation for
  the North American Land Data Assimilation System project phase 2 (NLDAS-2):
  1. Intercomparison and application of model products.
\newblock \emph{Journal of Geophysical Research: Atmospheres}, 117(D3).

\bibitem[{Xingjian et~al.(2015)Xingjian, Chen, Wang, Yeung, Wong, and
  Woo}]{Xingjian2015}
Xingjian, S.; Chen, Z.; Wang, H.; Yeung, D.~Y.; Wong, W.~K.; and Woo, W.~C.
  2015.
\newblock {Convolutional LSTM network: A machine learning approach for
  precipitation nowcasting}.
\newblock In \emph{Advances in neural information processing systems},
  802--810.

\bibitem[{Yang et~al.(2017)Yang, Dong, Sun, Lima, Mu, and Wang}]{yang2017cfcc}
Yang, Y.; Dong, J.; Sun, X.; Lima, E.; Mu, Q.; and Wang, X. 2017.
\newblock A CFCC-LSTM model for sea surface temperature prediction.
\newblock \emph{IEEE Geoscience and Remote Sensing Letters}, 15(2): 207--211.

\bibitem[{Zhang, Wang, and Wang(2002)}]{zhang2002precision}
Zhang, N.; Wang, M.; and Wang, N. 2002.
\newblock Precision agriculture—a worldwide overview.
\newblock \emph{Computers and electronics in agriculture}, 36(2-3): 113--132.

\bibitem[{Zhao et~al.(2020)Zhao, Song, Zhang, Liu, Wang, Lin, Deng, and
  Li}]{8809901}
Zhao, L.; Song, Y.; Zhang, C.; Liu, Y.; Wang, P.; Lin, T.; Deng, M.; and Li, H.
  2020.
\newblock T-GCN: A Temporal Graph Convolutional Network for Traffic Prediction.
\newblock \emph{IEEE Transactions on Intelligent Transportation Systems},
  21(9): 3848--3858.

\end{thebibliography}

\end{document}


\section{Supplementary File}

\section{Data}
Table \ref{tab:feat-descrip} contains the detailed description of features available at each node of the graph, in addition to soil moisture, for soil moisture forecasting. 
\begin{table}[hbtp!]
    \centering
    \begin{tabular}{|l|c|l|}
    \hline
    \multicolumn{1}{|c|}{\textbf{Name}} & \textbf{Units} & \multicolumn{1}{c|}{\textbf{Description}} \\ \hline
    AREA & $km^2$ & Drainage area of the HRU. \\ \hline
    PRECIP & $\text{mm} H_2O$ & \begin{tabular}[c]{@{}l@{}}Total amount of precipitation \\ falling on the HRU during \\ time step.\end{tabular} \\ \hline
    ET & $\text{mm} H_2O$ & \begin{tabular}[c]{@{}l@{}}Evapotranspiration from \\ the HRU during the time step.\end{tabular} \\ \hline
    PERC & $\text{mm} H_2O$ & \begin{tabular}[c]{@{}l@{}}Water that percolates past the\\ root zone during the time step.\end{tabular} \\ \hline
    GW-RCHG & $\text{mm} H_2O$ & \begin{tabular}[c]{@{}l@{}}Recharge entering aquifers \\ during time step.\end{tabular} \\ \hline
    DA-RCHG &  & Deep aquifer recharge. \\ \hline
    SOLAR & $MJ/m^2$ & Average daily solar radiation. \\ \hline
    SOL-TMP & °C & \begin{tabular}[c]{@{}l@{}}Average soil temperature of \\ first soil layer for time period.\end{tabular} \\ \hline
    DAILYCN & - & \begin{tabular}[c]{@{}l@{}}Average curve number for \\ time period.\end{tabular} \\ \hline
    TMP-AV & °C & Average daily air temperature. \\ \hline
    REVAP & $\text{mm} H_2O$ & \begin{tabular}[c]{@{}l@{}}Water in the shallow aquifer \\ returning to the root zone.\end{tabular} \\ \hline
    SA-IRR & $\text{mm} H_2O$ & Irrigation from shallow aquifer. \\ \hline
    DA-IRR & $\text{mm} H_2O$ & Irrigation from deep aquifer. \\ \hline
    SA-ST & $\text{mm} H_2O$ & Shallow aquifer storage. \\ \hline
    DA-ST & $\text{mm} H_2O$ & Deep aquifer storage. \\ \hline
    WYLD & $\text{mm} H_2O$ & \begin{tabular}[c]{@{}l@{}}Total amount of water leaving\\ the HRU and entering main \\ channel during the time step.\end{tabular} \\ \hline
    \end{tabular}
    \caption{List of features used and their description.}
    \label{tab:feat-descrip}
\end{table}

\section{Clustering}
\begin{figure}[hbtp!]
    \centering
    \includegraphics[width=0.47\textwidth]{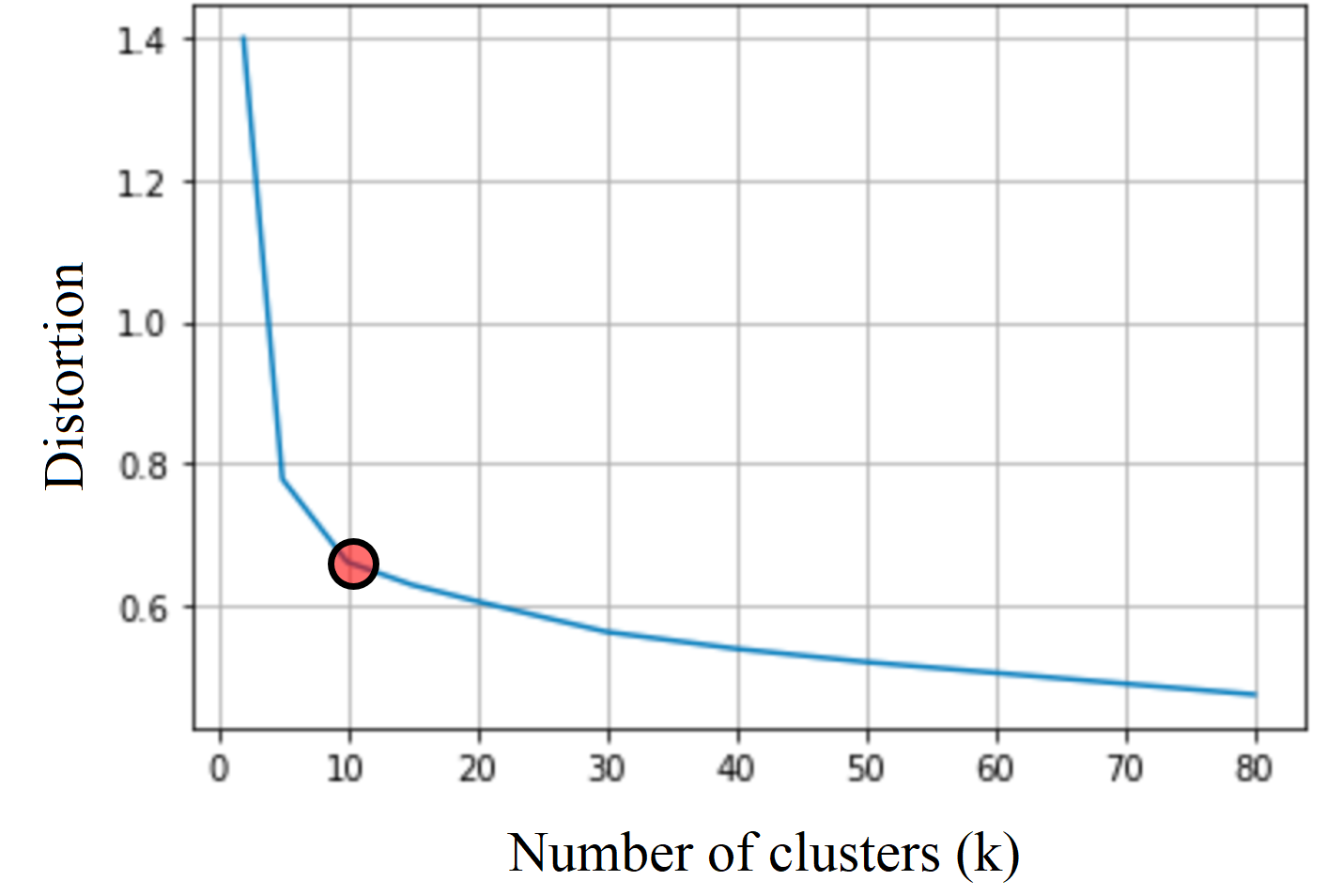}
    \caption{Elbow test for estimating number of clusters.}
    \label{fig:elbow}
\end{figure} 
The elbow method runs k-means clustering for a range of values for k and then for each value of k computes the distortion score for all clusters. The distortion score is the sum of square distances of each point from its assigned center.
\begin{figure}[hbtp!]
    \centering
    \includegraphics[width=0.46\textwidth]{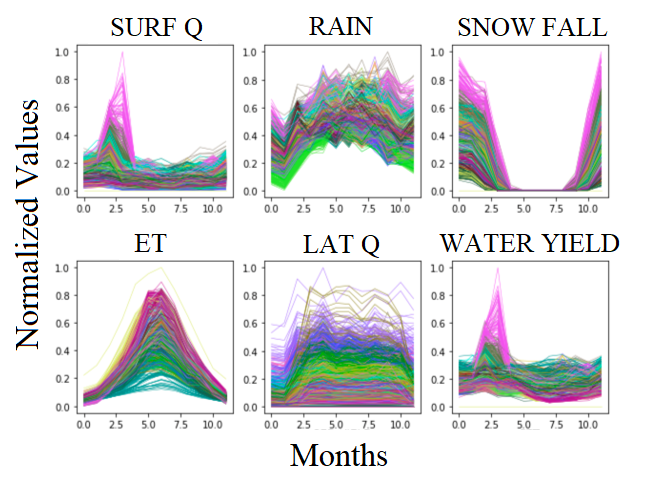}
    \caption{Clustering results f.}
    \label{fig:cluster}
\end{figure} %

A subset of features detailed in Table \ref{tab:cl_feat} is used to cluster the hydrological response units. Each feature time series consists of average annual values for each month. The features values are further normalized by the minimum and maximum values such that the range of values lies between 0 and 1. 
\begin{table}[hbtp!]
    \centering
    \begin{tabular}{|l|l|}
    \hline
    \textbf{Name} & \textbf{Description} \\ \hline
    SURF Q & \begin{tabular}[c]{@{}l@{}}Average annual \\ surface runoff in month\end{tabular} \\ \hline
    RAIN & \begin{tabular}[c]{@{}l@{}}Average annual \\ precipitation in month\end{tabular} \\ \hline
    \begin{tabular}[c]{@{}l@{}}SNOW \\ FALL\end{tabular} & \begin{tabular}[c]{@{}l@{}}Average annual \\ freezing rain in month\end{tabular} \\ \hline
    ET & \begin{tabular}[c]{@{}l@{}}Average annual \\ actual evapotranspiration in month\end{tabular} \\ \hline
    LAT Q & \begin{tabular}[c]{@{}l@{}}Average annual \\ lateral waterflow in month\end{tabular} \\ \hline
    \begin{tabular}[c]{@{}l@{}}WATER \\ YIELD\end{tabular} & \begin{tabular}[c]{@{}l@{}}Average annual \\ water yield in month\end{tabular} \\ \hline
    \end{tabular}
    \caption{Features used to for clustering and their description, all units are in mm.}
    \label{tab:cl_feat}
\end{table} 

\subsubsection{Libraries used}
\begin{enumerate}
    \item NumPy 1.21.5
    \item SciPy 1.7.3
    \item PTorch 1.11
    \item PyTorch Geometric Temporal
    \item Tslearn 0.5.2
\end{enumerate}

\begin{table*}
    \centering
        \begin{tabular}{|l|l|l|}
        \hline
        \textbf{Input Dataset} & \textbf{Source} & \textbf{Specifications} \\ \hline
        Watershed Boundaries & National Hydrography Dataset Plus 2.0 (NHDPlus) & Scale: HUC12 and HUC14 \\ \hline
        Elevation & \begin{tabular}[c]{@{}l@{}}USGS National Elevation Dataset (NED)- \\ Digital Elevation Model (DEM)\end{tabular} & \begin{tabular}[c]{@{}l@{}}Resolution: 30-m\\ Year: 2019\end{tabular} \\ \hline
        Stream Network & NHDPlus 2.0 & Year: 2019 \\ \hline
        Climate & \begin{tabular}[c]{@{}l@{}}Parameter-elevation Regressions on \\ Independent Slopes Model (PRISM) 2.0\end{tabular} & \begin{tabular}[c]{@{}l@{}}Period: 1981-2020 (Gridded)\\ Resolution: $\sim$4km\\ Scale: Monthly\end{tabular} \\ \hline
        Land Use (agricultural) & \begin{tabular}[c]{@{}l@{}}United States Department of Agriculture (USDA) \\ National Agricultural Statistics Service (NASS) \\ Cropland Data Layer (CDL)\end{tabular} & Year: 2018 \\ \hline
        \begin{tabular}[c]{@{}l@{}}Land Use\\ (non-agricultural)\end{tabular} & National Land Cover Database (NLCD) & Year: 2016 \\ \hline
        Soil & \begin{tabular}[c]{@{}l@{}}USDA Natural Resources Conservation Service (NRCS) \\ Soil Survey Geographic Data (SSURGO)\end{tabular} & \begin{tabular}[c]{@{}l@{}}Scale: County level\\ Year: 2019\end{tabular} \\ \hline
        \end{tabular}
    \caption{HAWQS v2.0 Input Data for SWAT simulation}
    \label{tab:hawqs}
\end{table*}